\def\year{2021}\relax
\documentclass[letterpaper]{article} 
\usepackage{aaai21}  
\usepackage{times}  
\usepackage{helvet} 
\usepackage{courier}  
\usepackage[hyphens]{url}  
\usepackage{graphicx} 
\usepackage{natbib}  
\usepackage{caption} 
\usepackage{color}
\usepackage{overpic}
\usepackage{verbatim}
\usepackage{graphicx}
\usepackage{wrapfig}
\usepackage{amsmath}
\usepackage{amssymb}
\usepackage{booktabs}
\usepackage{diagbox}
\usepackage{multirow}
\usepackage{arydshln}
\usepackage[switch]{lineno}

\usepackage{algorithmicx,algorithm}
\usepackage[noend]{algpseudocode}
\usepackage[dvipsnames]{xcolor}

\usepackage{algorithmicx,algorithm}
\title{Edge-competing Pathological Liver Vessel Segmentation with Limited Labels}
\author{
    Zunlei Feng\textsuperscript{\rm 1,\rm 5\#}, Zhonghua Wang\textsuperscript{\rm 1\#}, Xinchao Wang\textsuperscript{\rm 2}, Xiuming Zhang\textsuperscript{\rm 1}, Lechao Cheng\textsuperscript{\rm 3}, Jie Lei\textsuperscript{\rm 4},
    Yuexuan Wang\textsuperscript{\rm 1}\thanks{Corresponding author. \textsuperscript{\#}Both the authors have equal contribution to this work.},
     Mingli Song\textsuperscript{\rm 1,\rm 5}\\
}
\affiliations{
    \textsuperscript{\rm 1}Zhejiang University\\
    \textsuperscript{\rm 2}Stevens Institute of Technology\\
    \textsuperscript{\rm 3}Zhejiang Lab\\
    \textsuperscript{\rm 4}Zhejiang University of Technology\\
    \textsuperscript{\rm 5}Alibaba-Zhejiang University Joint Research Institute of Frontier Technologies\\

    \{zunleifeng,wang97zh,1508056,amywang,brooksong\}@zju.edu.cn, xinchao.wang@stevens.edu, chenglc@zhejianglab.com, jasonlei@zjut.edu.cn
    \vspace{1.2em}
}
\begin{document}

\def\mathbi#1{\textbf{\em #1}}

\maketitle

\begin{abstract}
The microvascular invasion (MVI) is a major prognostic factor in hepatocellular carcinoma,
which is one of the malignant tumors with the highest mortality rate.
The diagnosis of MVI needs discovering the vessels that contain hepatocellular carcinoma cells and counting their number in each vessel,
which depends heavily on experiences
of the doctor, is largely subjective and  time-consuming.
However, there is no algorithm as yet tailored for
the MVI detection from  pathological images.
This paper collects the first pathological liver image dataset
containing $522$ whole slide images with labels of vessels,
MVI, and hepatocellular carcinoma grades.
The first and essential step for the
automatic diagnosis of MVI is the accurate segmentation of vessels.
The unique characteristics of
 pathological liver images,
such as super-large size, multi-scale vessel, and blurred vessel edges,
make the accurate vessel segmentation challenging.
Based on the collected dataset, we propose an Edge-competing
Vessel Segmentation Network (EVS-Net), which contains a segmentation
network and two edge segmentation discriminators.
The segmentation network, combined with an edge-aware self-supervision mechanism,
is devised to conduct vessel segmentation with limited labeled patches.
Meanwhile, two discriminators are introduced to distinguish
whether the segmented vessel and background contain residual
features in an adversarial manner.
In the training stage, two discriminators are devised to
compete for the predicted  position of edges.
Exhaustive experiments demonstrate that, with only limited
labeled patches, EVS-Net achieves a close performance of fully
supervised methods, which provides a convenient tool for
the pathological liver vessel segmentation.
Code is publicly available at \url{https://github.com/zju-vipa/EVS-Net}.

\end{abstract}

\section{Introduction}
Liver cancer is one of the malignant tumors with the highest mortality rate,
which witnesses an increasing number of cases
in recent years ~\cite{d2017management}.
Hepatocellular carcinoma represents the most frequent primary liver tumor,
taking up about $90\%$ of all liver cancer cases ~\cite{perumpail2017clinical}.
The microvascular invasion (MVI) is a major prognostic factor in hepatocellular carcinoma, which significantly affects the choice of treatment and the overall survival of patients~\cite{sumie2008microvascular,rodriguezperalvarez2013a}.
However, the current MVI detection heavily depends on the experiences
of doctors, which is a largely
subjective and time-consuming to diagnose.
Existing machine learning algorithms for predicting MVI
mainly focused on computed tomography, magnetic resonance imaging, and positron emission tomography~\cite{cuccurullo2018microvascular}, yet ignore pathological images.
Even to data, there is no  learning-based method  yet for
automatically detecting MVI in pathological liver images.

\begin{figure}[!t]
\centering
\includegraphics[scale =0.34]{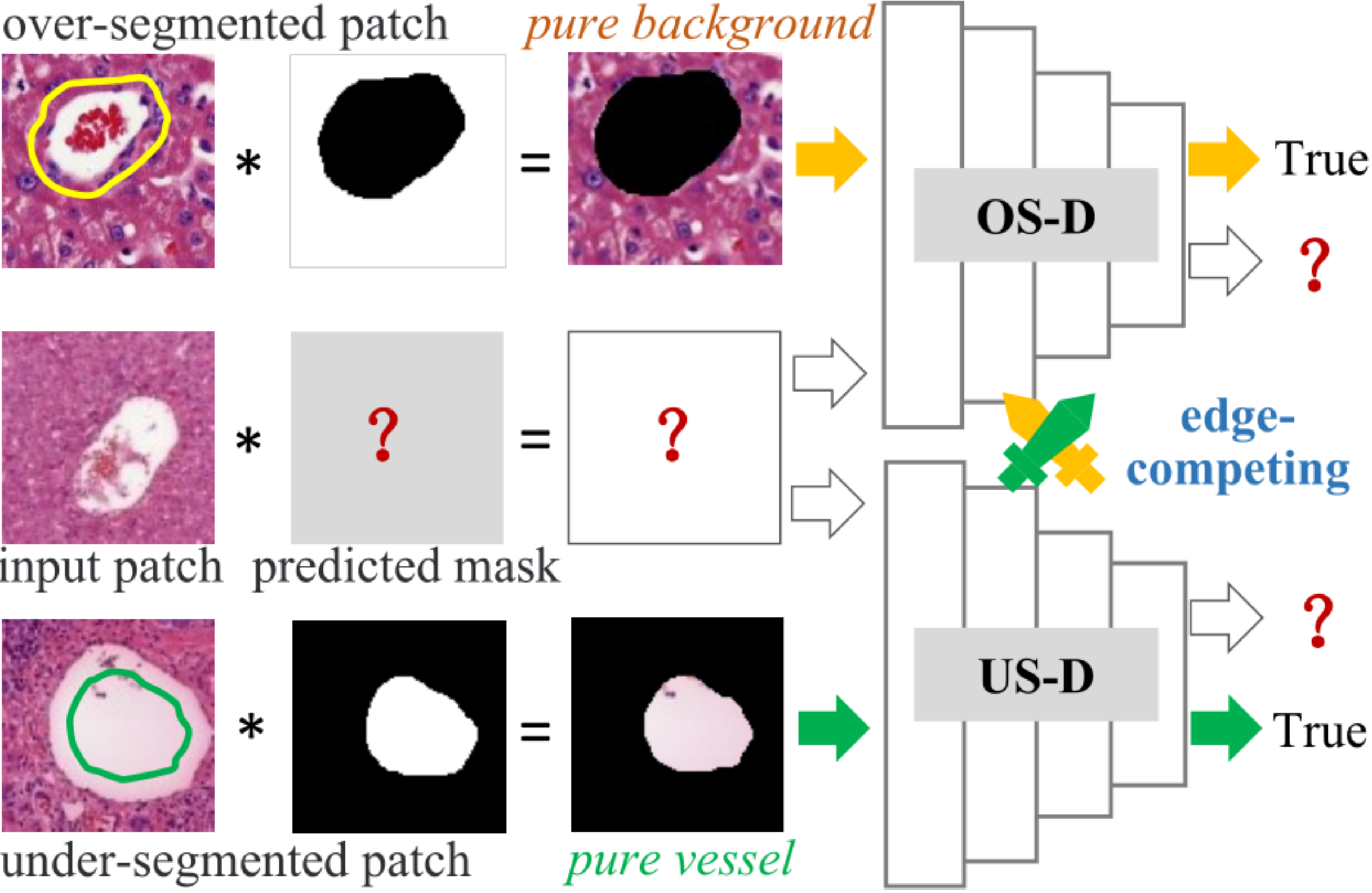}
\caption{The overview of the edge-competing mechanism.
The Over-Segmentation Discriminator (OS-D) and  Under-Segmentation Discriminator (US-D) are devised for distinguishing whether the segmented background and segmented vessel contain residual features.
The over-segmented and under-segmented patches are generated with different erosion and dilation operations following binarization processing.
Only when the predicted mask
depicts the precise segmentation
will the  true criterion
of the two discriminators
be satisfied
simultaneously for an input patch.}
\vspace{-1em}
\label{fig:overview}
\end{figure}

As deep learning approaches have achieved promising results in the medical image analysis area~\cite{ker2018deep,altaf2019going},
we expect they have the potential to
be deployed in the pathological liver vessel segmentation.
There are many deep learning based segmentation methods for general natural images~\cite{minaee2020image}
and medical images~\cite{ronneberger2015u-net:,xue2018segan:,Dai2019Transfer}.
Due to the specific attributes of different images, however,
existing methods cannot be directly applied to the
pathological liver vessel segmentation.
Moreover, the success of deep learning are highly based on a large number of annotated samples;
yet for pathological liver vessel segmentation,
there was not a dedicated dataset available,
as it is extremely  tedious, time-consuming, and expensive
to annotate those vessels accurately.

\begin{figure*}[!t]
\centering
\vspace{-0.5em}
\includegraphics[scale =0.585]{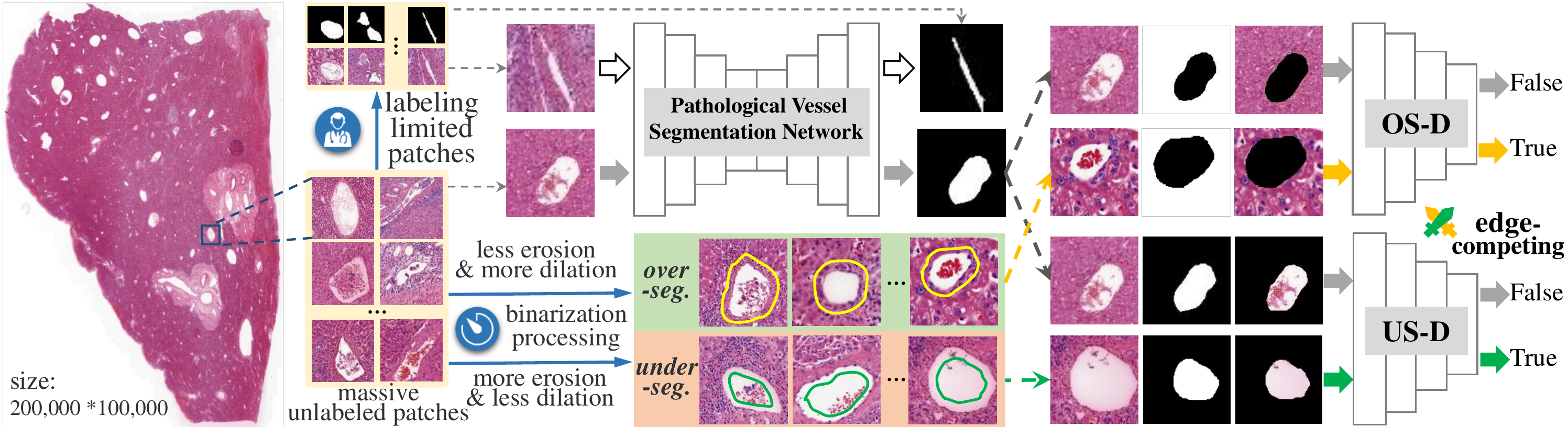}
\caption{The flow diagram of pathological liver vessel segmentation.
For massive unlabeled patches cut from the whole slide pathological image, the doctor only needs to label limited patches.
Meanwhile, some under-segmented (\emph{over-seg.}) patches and under-segmented (\emph{under-seg.}) patches are generated
through different erosion and dilation operations  following binarization processing.
Then, the EVS-Net is proposed for segmenting the vessels in the patches based on the above limited labels, over-segmented and under-segmented patches.
The EVS-Net contains a Pathological Vessel Segmentation Network (PVSN), an over-segmentation discriminator (OS-D), and an under-segmentation discriminator (US-D).
The PVSN is first initialized by training with the limited labels.
With the over-segmented and under-segmented patches, the two discriminators are trained for distinguishing whether the segmented vessels and background contain residual features. In the training stage, for an unlabeled patch, only when the predicted mask is the perfect segmentation will the true criterion of the two discriminators be satisfied simultaneously, which facilitates the PVSN segmenting more accurately.
}
\vspace{-0.5em}
\label{fig:framework}
\end{figure*}

In this paper, we first collect a pathological liver
image dataset containing $522$ whole slide images with
labels of vessels, MVI, and hepatocellular carcinoma grades.
It can readily serve as a benchmark for the research on the
analysis of pathological liver images.
Fig.~\ref{fig:framework} shows some cut patches and the whole
slide pathological image, which has a size of $200,000 * 100,000$.
For the patches, the coarse edges of most vessels can be obtained by binarization processing. For the vessels containing hepatocellular carcinoma cells, binarization processing will fail to segment those patches.
However, combining with the erosion and dilation operation, binarization processing may easily produce some over-segmented or under-segmented vessels.

Based on such findings, we devise an edge-competing mechanism to assist the vessel segmentation, for which
the overview
is shown in Fig.~\ref{fig:overview}.
With the easily obtained over-segmented patches as the positive training samples,
the over-segmentation discriminator is devised for distinguishing whether the segmented background contains  residual features of  vessels.
Similarly, the under-segmentation discriminator is  devised for distinguishing whether the segmented vessels contain  residual features of the background.
For an input patch, only when the predicted mask depicts the perfect segmentation will the true criterion of the two discriminators
be satisfied simultaneously.
It will effectively relieve the dependence of annotation by incorporating the edge-competing mechanism into the vessel segmentation framework.

Besides, from the whole slide image in Fig.~\ref{fig:framework}, we can see that the pathological liver images have some unique attributes, such as super-large sample size, multi-scale vessel, blurred vessel edge,
which is challenging for the accurate vessel segmentation.
Given such specific characteristics of pathological liver images,
we propose the Edge-competing Vessel Segmentation Network (EVS-Net), which is shown in Fig.~\ref{fig:framework}.
For the massive unlabeled patches, the doctor only needs to label limited patches
as guidance.
For the rest of the unlabeled patches, the binarization processing combined with erosion and dilation will generate
many over-segmented and under-segmented patches.

As shown in Fig.~\ref{fig:framework}, the proposed EVS-Net is composed of a Pathological Vessel Segmentation Network (PVSN), an Over-Segmentation Discriminator (OS-D), and an Under-Segmentation Discriminator (US-D).
In the training stage,
the limited labeled  patches will supervise the PVSN to gain initial segmentation ability for the pathological vessel patches.
The edge-competing mechanism is then incorporated into EVS-Net for assisting the training of the segmentation network in an adversarial manner.

To further strengthen  consistency
of the segmentation
on massive unlabeled patches,
an edge-aware self-supervision module is
introduced to generate the self-supervision information for training PVSN.
The core idea of edge-aware self-supervision is that,
for the same image and segmentation network, the predicted mask of the \emph{transformed image} should be equal to the \emph{transformed mask} predicted by the network with the original image as input.

Our contribution can be summarized as follows.
Firstly, we collect a pathological liver image dataset
that contains $522$ whole slide images with labels of vessels,
MVI, and hepatocellular carcinoma grades.
It can readily serve as  a benchmark for the analysis of hepatocellular carcinoma.
Next, the proposed segmentation network combines an edge-aware self-supervision module and two segmentation discriminators to
efficiently solve the challenges in the pathological vessel segmentation.
Furthermore, the edge-competing mechanism between two discriminators is firstly proposed for learning accurate segmentation
with over-segmented and under-segmented patches, which  effectively relieves the dependence on a large number of annotations.
Finally, with limited labels, the EVS-Net achieves results on par with fully supervised ones,
providing a handy tool for the pathological liver vessel segmentation.

\section{Related Works}
To our knowledge, there are no vessel
segmentation methods for pathological images until now.
Existing methods, which are related
to pathological images, mainly focus on
the pathological grading of various carcinomas~\cite{yang2019noninvasive,jiao2020a,liao2020d}.
Here, we give the related segmentation works for general natural images and medical images from two perspectives.


\textbf{Edge-aware Segmentation}.
For edge-aware semantic segmentation, the commonly adopted framework is a two-branch network that simultaneously predicts segmentation maps and edges, where different constraints \citep{bertasius2016semantic,chen2016semantic,heng2017fusionNet:,yu2018learning,Yin_2018_ECCV,Ye_2019_CVPR,takikawa2019gated-scnn:} are devised for strengthening the segmentation results with the predicted edges. Unlike predicting the edge directly,
\citet{hayder2017boundary-aware} proposed predicting pixels' distance to the edge of object and post-processed the distance map into the final segmented results.
 ~\citet{khoreva2017simple} proposed edge-aware filtering to improve object delineation.
\citet{zhang2017global-residual} proposed a local edge refinement network to learn the position-adaptive propagation coefficients so that local contextual information from neighbors can be optimally captured for refining object edges.
\citet{qin2019basnet:} adopted the patch-level SSIM loss~\citep{wang2003multiscale} to assign higher weights to the edges.
\citet{peng2017large} proposed an edge refinement block to improve the localization performance near the object edges.
Unlike the above methods,
two edge-competing discriminators are devised for distinguishing whether the segmented vessels and background contain residual features.
The over-segmented and under-segmented patches are adopted to train the two discriminators, focusing on the outer and inner area
of edges.

\textbf{GAN based Segmentation} contains two categories: composition fidelity based methods~\cite{ostyakov2018seigan:,remez2018learning,chen2019unsupervised,QiuECCV_2020} and mask distribution-based methods ~\cite{luc2016semantic,han2017transferring,arbelle2018microscopy,hung2018adversarial,Ye_2020_CVPR}.
The composition fidelity based methods~\cite{ostyakov2018seigan:,remez2018learning,chen2019unsupervised} adopted the discriminator to distinguish the fidelity of natural images and the composited images, which is composition of the segmented objects and some background images.
For the mask distribution-based methods, Luc~\emph{et al.}~\cite{luc2016semantic} adopted the adversarial optimization between segmented results and GT mask to train the segmentation network, which is the first GAN based semantic segmentation network. Next, some researchers~\cite{han2017transferring,arbelle2018microscopy,xue2018segan:} applied the same adversarial framework on the medical image segmentation task. Furthermore, Souly~\emph{et al.} \cite{souly2017semi} and Hung~\emph{et al.} \cite{hung2018adversarial} adopted the adversarial strategy to train the discriminator output class confidence maps.
Unlike the above existing GAN-based methods, we devised two-discriminator architecture,
where the two discriminators will compete the vessel edge's position of the unlabeled patch in the adversarial training stage.
In addition, the true inputs of the two discriminators are over-segmented and under-segmented patches.

\section{Proposed Method}
In the real diagnostic scenario, the assisting diagnosis tool of MVI should be objective, accurate, and convenient. The less time and labor cost, the better.
Based on the real requirement of diagnosis, we introduce a pathological liver vessel segmentation method that can achieve applicable performance with limited labels.
The proposed method contains two parts:
the pathological patches pre-processing and pathological vessel segmentation.
In the former part, the doctor only needs to label limited patches.
Some over/under-segmented patches are generated with different erosion and dilation operations following binarization processing. In the latter part,
the EVS-Net is devised for segmenting the pathological liver vessels based on the above limited labeled patches, over-segmented and under-segmented patches.

\subsection{Pathological Patch Preprocessing}
From Fig.~\ref{fig:framework}, we can see that the pathological liver image has a super-large size and multi-scale vessel.
Meanwhile, the areas of vessels and background are unbalanced. To get the useful training samples, we first locate the vessels' general positions according to the white color.
Then, three sizes are adopted to cut patches from the whole slide image around the located positions.

For the massive cut patch set $\mathbi{S}=\{\mathbi{x}_1,\mathbi{x}_2,\mathbi{x}_3,...,\mathbi{x}_N\}$,
the doctor only needs to label limited patches $\mathbb{S}=\{(\mathbi{x}_1,\mathbi{m}_1), (\mathbi{x}_2,\mathbi{m}_2),(\mathbi{x}_3,\mathbi{m}_3), ..., (\mathbi{x}_K,\mathbi{m}_K)\}$.
Next, for a patch $\mathbi{x}\in \mathbi{S}$,
the under-segmented mask $\overline{\mathbi{m}}$  of patch $\mathbi{x}$  can be calculated as follows:
\begin{equation}\label{eq1}
\overline{\mathbi{m}}=[\mathbf{1}] - T_{ostu}(\mathbi{x})\ominus \mathfrak{S}_{r1} \oplus \mathfrak{S}_{r2}, r1>r2,
\end{equation}
where, $[\mathbf{1}]$ denotes the unit matrix that has the same size of the mask, $\mathcal{T}_{ostu}$ denotes the OTSU binarization algorithm~\cite{otsu1979a},
$\mathfrak{S}$ denotes the disk structuring element.
$\ominus$ denotes the erosion operation by structure $\mathfrak{S}_{r1}$ with radius $r1$,
$\oplus$ denotes the dilation operation by structure $\mathfrak{S}_{r2}$ with radius $r2$.
The $r1$ and $r2$ are integers.
The obtained $\overline{\mathbi{m}}$ is the mask of background.
The over-segmented patches constitute the over-segmented image set $\mathbb{O}=\{(\mathbi{x}_1,\overline{\mathbi{m}}_1), (\mathbi{x}_2,\overline{\mathbi{m}}_2),(\mathbi{x}_3,\overline{\mathbi{m}}_3), ..., (\mathbi{x}_N,\overline{\mathbi{m}}_N)\}$.

Similarly, the under-segmented image set $\mathbb{U}=\{(\mathbi{x}_1,\underline{\mathbi{m}}_1), (\mathbi{x}_2,\underline{\mathbi{m}}_2),(\mathbi{x}_3,\underline{\mathbi{m}}_3), ..., (\mathbi{x}_N,\underline{\mathbi{m}}_N)\}$ is calculated as follows:
\begin{equation}\label{eq2}
\underline{\mathbi{m}}=T_{ostu}(\mathbi{x})\ominus \mathfrak{S}_{r1} \oplus \mathfrak{S}_{r2}, r1<r2.
\end{equation}

\subsection{Pathological Vessel Segmentation}
As shown in Fig.~\ref{fig:framework}, the proposed EVS-Net is composed of a segmentation network and two discriminators.
The segmentation network is trained with supervision loss on the labeled limited patches
and an edge-aware self-supervision loss on unlabeled patches.
The two discriminators are trained on over-segmented and under-segmented patches for distinguishing whether the segmented background and vessel contain residual features, respectively.
In the adversarial training stage, the edge competition between two discriminators will facilitate the vessels' accurate segmentation.

\subsubsection{Segmentation Network.}
In the EVS-Net, the segmentation network $\mathcal{F}_{\theta}$ is designed to be an encoder-decoder architecture. With a labeled patch $\mathbi{x}\in \mathbb{S}$, the segmented result $\ddot{\mathbi{m}}=\mathcal{F}_{\theta}(\mathbi{x})$ is expected to approximate the GT mask $\mathbi{m}$, which can be achieved by minimizing the pixel-wise two-class Dice loss $\mathbf{\mathcal{L}_{dic}}$:
\begin{equation}\label{eq1}
\mathbf{\mathcal{L}_{dic}}=1-\frac{2|\ddot{\mathbi{m}} \cap \mathbi{m}|+\zeta}{|\ddot{\mathbi{m}}|+|\mathbi{m}|+\zeta},
\end{equation}
where, $\zeta$ is the Laplace smoothing parameter for preventing zero error and reducing overfitting.
For the labeled samples, we augment samples' diversity through compositing background.
The composited background is generated by adding other pure background and normalizing the new values into the normal range.

\emph{Edge-aware Self-supervision.} To reduce the dependence on massive labeled samples,
inspired by ~\citet{wang2019self-supervised}, we introduce an edge-aware self-supervision strategy, which can strengthen the edge consistency of vessels.
The core idea is that the predicted mask of a transformed image should be equal to the transformed mask predicted by the network with the original image as input.
Formally, for the robust segmentation network, given an affine transformation matrix $\mathbi{M}$, segmented result $\mathcal{F}_{\theta}(\mathbi{M}\mathbi{x})$ of the affine transformed image should be consistent with the affine transformed mask  $\mathbi{M}\mathcal{F}_{\theta}(\mathbi{x})$ in the following way: $\mathcal{F}_{\theta}(\mathbi{M}\mathbi{x})=\mathbi{M}\mathcal{F}_{\theta}(\mathbi{x})$. Furthermore, we obtain the edge neighborhood weight map $\mathbi{w}$ as follows:
\begin{equation}\label{eq2}
\mathbi{w}=\ddot{\mathbi{m}} \oplus \mathfrak{S}_{r3} -\ddot{\mathbi{m}} \ominus \mathfrak{S}_{r3},
\end{equation}
where, $r3$ denotes the radius of the structure $\mathfrak{S}$.
With the weight map $\mathbi{w}$, the edge-aware self-supervision loss $\mathbf{\mathcal{L}_{sel}}$ is defined as follows:
\begin{equation}\label{eq3}
\mathbf{\mathcal{L}_{sel}}=||\mathbi{w}'\mathcal{F}_{\theta}(\mathbi{M}\mathbi{x})-\mathbi{M}\{\mathbi{w}\mathcal{F}_{\theta}(\mathbi{x})\}||^2_2,
\mathbi{x} \in \mathbi{S},
\end{equation}
where, $\mathbi{w}'$ and $\mathbi{w}$ are the weight maps of the predicted masks $\mathcal{F}_{\theta}(\mathbi{M}\mathbi{x})$ and $\mathcal{F}_{\theta}(\mathbi{x})$, respectively.
The edge-aware self-supervision mechanism not only strengthens the edge consistency of the segmented vessels but also can eliminate the unreasonable holes in the predict masks. The root reason is that unreasonable holes will be assigned high weights.
\subsubsection{Edge-competing Discriminator.}
Based on the obtained under-segmented set $\mathbb{O}$ and under-segmented set $\mathbb{U}$,
we devised two discriminators for distinguishing whether the segmented patch is a pure vessel or a pure background.

\emph{Over-segmentation Discriminator.}
For the over-segmented patch pairs $(\mathbi{x},\overline{\mathbi{m}}) \in \mathbb{O}$,
the background $\mathbi{x}_b$ is computed using the following equation:
$\mathbi{x}_b=\overline{\mathbi{m}}*\mathbi{x}$,
where `$*$' denotes pixel-wise multiplication.
Then the concatenated triplet $\mathbi{I}_b=[\mathbi{x},\overline{\mathbi{m}},\mathbi{x}_b]$ is set as the positive sample of OS-D.
The true condition of OS-D is that the segmented background doesn't contain any residual vessels' features.
For the unlabeled patch $\mathbi{x} \in \mathbi{S}$,
the segmentation network predicts the mask of vessel $\ddot{\mathbi{m}}=\mathcal{F}_{\theta}(\mathbi{x})$.
The reversed  mask is calculated as follows:  $\ddot{\mathbi{m}}'=[\mathbf{1}]-\ddot{\mathbi{m}}$, where the $[\mathbf{1}]$
denotes the unit matrix that has the same size of the mask.
Next, the segmented background $\ddot{\mathbi{x}}_b$ is computed using the following equation:
$\ddot{\mathbi{x}}_b=\ddot{\mathbi{m}}'*\mathbi{x}$,
Then the concatenated triplet $\ddot{\mathbi{I}}_b=[\mathbi{x},\ddot{\mathbi{m}}',\ddot{\mathbi{x}}_b]$ is fed to the OS-D $\mathcal{D}^b_{\phi}$, which discriminates whether the segmented background $\ddot{\mathbi{x}}_b$ contains the vessels' residual features.
The $\ddot{\mathbi{I}}_b$ is regarded as a false triplet.
The adversarial optimization between the segmentation network and OS-D will constrain the segmented background predicted by the segmentation network
don't contain vessels' features.
The over-segmentation adversarial loss $\mathbf{\mathcal{L}^{ovr}_{adv}}$ is given as follows:
\begin{equation}\label{eq4}
\begin{split}
\mathbf{\mathcal{L}^{ovr}_{adv}}& = \mathop{\mathbb{E}}\limits_{\ddot{\mathbi{I}}_b \sim \ddot{\mathbb{P}}_b} [\mathcal{D}^b_{\phi}(\ddot{\mathbi{I}}_b)]
-\mathop{\mathbb{E}}\limits_{\mathbi{I}_b \sim \mathbb{P}_b} [\mathcal{D}^b_{\phi}(\mathbi{I}_b)]  \\
& +\lambda \mathop{\mathbb{E}}\limits_{\hat{\mathbi{I}}_b\sim \mathbb{P}_{\hat{\mathbi{I}}_b}} [(\|\nabla_{\hat{\mathbi{I}}_b} \mathcal{D}^{b}_{\phi}(\hat{\mathbi{I}}_b)\|_{2} - 1)^2],
\end{split}
\end{equation}
where, the $\ddot{\mathbb{P}}_b$, $\mathbb{P}_b$ are the generated background triplet distribution and under-segmented triplet distribution, respectively.
The $\hat{\mathbi{I}}_b$ is sampled uniformly along straight lines between pairs of points sampled from the generated background triplet distribution $\ddot{\mathbb{P}}_b$ and the under-segmented triplet distribution $\mathbb{P}_b$.
The $\hat{\mathbi{I}}_b=\varepsilon \mathbi{I}_b+(1-\varepsilon)\ddot{\mathbi{I}}_b$, where the $\varepsilon$ is a random number between $0$ and $1$.
The second term (gradient penalty) is firstly proposed in WGAN-GP~\cite{gulrajani2017improved}. The $\lambda$ is the gradient penalty coefficient.

\emph{Under-segmentation Discriminator.}
For the under-segmented patch pairs $(\mathbi{x},\underline{\mathbi{m}}) \in \mathbb{U}$,
the segmented vessel $\mathbi{x}_v$ is calculated with the following equation:
$\mathbi{x}_v=\underline{\mathbi{m}}*\mathbi{x}$.
Then the concatenated triplet $\mathbi{I}_v=[\mathbi{x},\underline{\mathbi{m}},\mathbi{x}_v]$ satisfies the true condition of the US-D
that the segmented vessel doesn't contain any residual background features.
For the unlabeled patch $\mathbi{x} \in \mathbi{S}$,
the concatenated triplet $\ddot{\mathbi{I}}_v=[\mathbi{x},\ddot{\mathbi{m}},\ddot{\mathbi{x}}_v]$ is fed to the US-D $\mathcal{D}^v_{\varphi}$, which distinguishes whether the segmented vessel $\ddot{\mathbi{x}}_v$ contains the background residual features.
The $\ddot{\mathbi{I}}_v$ is regarded as the false triplet for US-D.
The under-segmentation adversarial loss $\mathbf{\mathcal{L}^{udr}_{adv}}$ is given as follows:
\begin{equation}\label{eq4}
\begin{split}
\mathbf{\mathcal{L}^{udr}_{adv}}& = \mathop{\mathbb{E}}\limits_{\ddot{\mathbi{I}}_v \sim \ddot{\mathbb{P}}_v} [\mathcal{D}^v_{\varphi}(\ddot{\mathbi{I}}_v)]
-\mathop{\mathbb{E}}\limits_{\mathbi{I}_v \sim \mathbb{P}_v} [\mathcal{D}^v_{\varphi}(\mathbi{I}_v)]  \\
& +\lambda \mathop{\mathbb{E}}\limits_{\hat{\mathbi{I}}_v\sim \mathbb{P}_{\hat{\mathbi{I}}_v}} [(\|\nabla_{\hat{\mathbi{I}}_v} \mathcal{D}^{v}_{\varphi}(\hat{\mathbi{I}}_v)\|_{2} - 1)^2],
\end{split}
\end{equation}
where, the $\ddot{\mathbb{P}}_v$, $\mathbb{P}_v$ are the generated vessel triplet distribution and under-segmented triplet distribution, respectively.
The $\hat{\mathbi{I}}_v$ is sampled uniformly along straight lines between pairs of points sampled from the generated vessel triplet distribution $\ddot{\mathbb{P}}_v$ and the under-segmented triplet distribution $\mathbb{P}_v$.
The $\hat{\mathbi{I}}_v=\varepsilon \mathbi{I}_v+(1-\varepsilon)\ddot{\mathbi{I}}_v$, where the $\varepsilon$ is a random number between $0$ and $1$.
The optimization on $\mathbf{\mathcal{L}^{udr}_{adv}}$ will constrain the segmented vessel predicted by the segmentation network
don't contain background features.

\subsection{Complete Algorithm}
To sum up, the supervised Dice loss $\mathbf{\mathcal{L}_{dic}}$ is adopted to train the pathological vessel segmentation network with the limited labeled patches.
Then, edge-aware self-supervision is introduced to strengthen the edge consistency of the segmented mask with massive unlabeled patches.
For the over-segmented patches and under-segmented patches,
the edge adversarial losses $\mathbf{\mathcal{L}^{ovr}_{adv}}$ and $\mathbf{\mathcal{L}^{udr}_{adv}}$ are adopted to train the OS-D and US-D.
In the training stage, the two discriminators compete for the edge's position of the predicted mask, which will refine the segmented mask.
During training, we alternatively optimize the segmentation network $\mathcal{F}_{\theta}$ and two edge discriminators $ \mathcal{D}^b_{\phi},\mathcal{D}^v_{\varphi}$ using the randomly sampled samples from
unlabeled patch set $\mathbi{S}$, labeled patch set $\mathbb{S}$, under-segmented set $\mathbb{O}$, and under-segmented set $\mathbb{U}$. The complete algorithm is summarized in Algorithm~\ref{alg:alg1}.

\begin{algorithm}[!h]
\caption{The Training Algorithm for EVS-Net}
\begin{algorithmic}[1]
\renewcommand{\algorithmicrequire}{\textbf{Require:}}
\renewcommand{\algorithmicensure}{\textbf{Require:}}
\Require{ The gradient penalty coefficient $\lambda$, interval iteration number
$n_{critic}$, the batch size $T$, the Laplace smoothing parameter $\zeta$, Adam hyperparameters $\alpha,\beta_{1},\beta_{2}$, the balance parameters $\tau,\eta$ for $\mathcal{L}_{dic}$ and $\mathbf{\mathcal{L}_{sel}}$.}
\Require{ Initial critic parameters $\varphi,\phi$, initial segmentation networt parameters $\theta$. }
\While{$\theta$ has not converged}
\For{$i=1,...,n_{critic}$}
\For{$t=1,...,T$}
\State Sample $\mathbi{x} \in \mathbi{S}$, $(\mathbi{x},\overline{\mathbi{m}}) \in \mathbi{O}$, $(\mathbi{x},\underline{\mathbi{m}}) \in \mathbi{U}$,
\Statex\quad \quad \quad \quad \quad a random number $\varepsilon \sim [0,1]$.
\State Obtain real triplet $\mathbi{I}_b$ and $\mathbi{I}_v$.
\State Obtain false triplet $\ddot{\mathbi{I}}_b$ and $\ddot{\mathbi{I}}_v$.
\State $\mathbf{\mathcal{L}^{ovr}_{adv}}^{(t)}\!\leftarrow\!\mathcal{D}^b_{\phi}(\ddot{\mathbi{I}}_b)\!-\!\mathcal{D}^b_{\phi}(\mathbi{I}_b) $
\Statex \quad \qquad \quad \qquad \quad \qquad $+ \lambda(\|\nabla_{\hat{\mathbi{I}}_b} \mathcal{D}^{b}_{\phi}(\hat{\mathbi{I}}_b)\|_{2} - 1)^2$.
\State $\mathbf{\mathcal{L}^{udr}_{adv}}^{(t)} \leftarrow \mathcal{D}^v_{\varphi}(\ddot{\mathbi{I}}_v)- \mathcal{D}^v_{\varphi}(\mathbi{I}_v)$
\Statex \quad \qquad \quad \qquad \quad \qquad $+ \lambda(\|\nabla_{\hat{\mathbi{I}}_v} \mathcal{D}^{v}_{\varphi}(\hat{\mathbi{I}}_v)\|_{2} - 1)^2$.
\EndFor
\State $\phi \leftarrow Adam(\nabla_{\phi}\frac{1}{T} \sum^T_{t=1}\mathbf{\mathcal{L}^{ovr}_{adv}}^{(t)},\phi,\alpha,\beta_1,\beta_2)$.
\State $\varphi \leftarrow Adam(\nabla_{\varphi}\frac{1}{T} \sum^T_{t=1}\mathbf{\mathcal{L}^{udr}_{adv}}^{(t)},\varphi,\alpha,\beta_1,\beta_2)$.
\EndFor
\State Sample unlabeled batch $\{\mathbi{x}_t\}^T_{t=1}$  from $\mathbi{S}$  and
\Statex \quad \quad labeled batch $\{(\mathbi{x}_t, \mathbi{m}_t)\}^T_{t=1}$ from $\mathbb{S}$.
\State $\mathbf{\mathcal{L}_{dic}}\leftarrow 1-\frac{2|\mathcal{F}_{\theta}(\mathbi{x}) \cap \mathbi{m}|+\zeta}{|\mathcal{F}_{\theta}(\mathbi{x})|+|\mathbi{m}|+\zeta}$.
\State $\mathbf{\mathcal{L}_{sel}}\leftarrow ||\mathbi{w}'\mathcal{F}_{\theta}(\mathbi{M}\mathbi{x})-\mathbi{M}\{\mathbi{w}\mathcal{F}_{\theta}(\mathbi{x})\}||^2_2$.
\State $\theta \leftarrow Adam(\nabla_{\theta}\frac{1}{T} \sum^T_{t=1}\{\mathbf{\tau\mathcal{L}_{dic}}+ \eta\mathbf{\mathcal{L}_{sel}}$
\Statex \quad \quad \quad $-\mathcal{D}^{b}_{\phi}(\mathcal{F}_{\theta}(\mathbi{x}))-\mathcal{D}^{v}_{\varphi}(\mathcal{F}_{\theta}(\mathbi{x}))\}),\theta,\alpha,\beta_1,\beta_2)$.
\EndWhile
\State \Return Segmentation networt parameters $\theta$, critic parameters $\varphi,~\phi$.
\end{algorithmic}
\label{alg:alg1}
\end{algorithm}

\section{Experiments}
\textbf{Dataset.} The collected pathological liver image dataset contains $522$ whole slide images.
Each image size has about $200,000*100,000$.
 For each image, pathologists labeled the vessels' masks, MVI, and hepatocellular carcinoma grades.
 There are total $98,398$ vessels where $2,890$ vessels contain hepatocellular carcinoma cells.
 The sample number for levels one, two, three and four tumor are $52$, $245$, $189$, and $36$, respectively.
It can be a benchmark for the research on the analysis of pathological liver images.
For all images, we finally collect $180,000$ patches, which have three scales ($128*128$, $512*512$ and $1024*1024$).
All patches are resized into $128*128$. The train, validation, and test sample number are $144,000$, $18,000$, and $18,000$, respectively.

\textbf{Network architecture and parameters.} The segmentation network we adopted is the DeeplabV3+ (backbone: resnet50 )~\cite{chen2017rethinking}. In the experiment, the parameters are set as follows: $\zeta=1, \tau =1,\eta =2, \lambda = 10$, $n_{critic}=5$, the batch size $T =64$, Adam hyperparameters for two discriminators $\alpha =0.0001,\beta_{1}=0,\beta_{2}=0.9$.
The radius $r1$ and $r2$ are random values between $5$ and $30$. The radius $r3$ equals to $15$.
The learning rate for the segmentation network and two discriminators are all set to be  $1e^{-4}$.

\textbf{Metric.} The metrics we adopted include Pixel Accuracy (PA), Mean Pixel Accuracy (MPA), Mean Intersection over Union (MIoU), Frequency Weighted Intersection over Union (FWIoU), and Dice.

\begin{table*}[!t]
\resizebox{\textwidth}{!}{
\centering
\begin{tabular}{c| c cc cc ccccccc c}
\toprule
 &\textbf{Type}
    & \multicolumn{2}{c}{\emph{Unsupervised}}
    & \multicolumn{2}{c}{\emph{Weakly Supervised}}
     & \multicolumn{2}{c}{\emph{Edge-aware}}
    & \multicolumn{5}{c}{\emph{Fully Supervised}}
    & \emph{Ours}\\
   \cmidrule(r){2-2} \cmidrule(r){3-4}  \cmidrule(r){5-6} \cmidrule(r){7-8} \cmidrule(r){9-13} \cmidrule(r){14-14}
&\textbf{Method}
    & \textbf{ReDO}
    & \textbf{CAC}
    & \textbf{ALSSS/ 500}
    & \textbf{USSS/ 500}
    & \textbf{G-SCNN}
    & \textbf{BFP}
    & \textbf{Unet}
    & \textbf{FPN}
    & \textbf{PSPNet}
    & \textbf{PAN}
    & \textbf{Deeplab}
    & \textbf{E(100)} \\

\midrule
\multirow{4}{*}{\rotatebox{90}{All patch}}
&\textbf{MPA}        &87.89 &84.32    &62.92/ 91.84 &89.88/ 94.93   &97.29 &96.83 &96.45   &98.56 &98.29  &98.44 &\underline{98.82}  &\textbf{96.59}    \\
&\textbf{MIOU}       &82.20 &78.35    &55.41/ 89.40 &84.57/ 92.35   &97.28 &96.04 &92.76  &97.40 &96.86  &97.11 &\underline{97.79}   &\textbf{94.38}   \\
&\textbf{FWIoU}      &89.34 &80.57    &74.45/ 93.77 &90.75/ 95.46   &95.89 &96.32 &95.64  &98.46 &98.13  &98.29 &\underline{98.69}   &\textbf{96.67}    \\
&\textbf{DICE}       &96.56 &88.32    &91.99/ 98.08 &97.02/ 98.58   &98.76 &98.21 &98.62  &99.52  &99.42  &99.47 &\underline{99.60}   &\textbf{98.96}    \\
\cdashline{1-14}[0.8pt/2pt]
\multirow{4}{*}{\rotatebox{90}{MVI patch}}
&\textbf{MPA}        &81.85 &78.32     &56.09/ 77.71 &80.61/ 89.34   &95.87 &95.43 &89.14  &97.55  &96.61  &97.17 &\underline{97.97}  &\textbf{90.85}   \\
&\textbf{MIOU}       &74.06 &69.87     &48.21/ 72.43 &73.82/ 84.83   &92.85 &94.72 &81.77  &95.60  &94.45  &94.82 &\underline{95.84} &\textbf{86.19}    \\
&\textbf{FWIoU}      &84.75 &79.10     &71.55/ 84.44 &84.81/ 91.24   &95.90 &94.53 &89.16 &97.48  &96.80  &97.00 &\underline{97.60}  &\textbf{91.98}  \\
&\textbf{DICE}       &94.95 &85.32     &91.28/ 95.19 &95.09/ 97.25   &94.83 &97.53 &96.42  &99.23  &99.01  &99.07 &\underline{99.29}  & \textbf{97.46}    \\
\bottomrule
\end{tabular}
}
\vspace{-0.6em}
\caption{The quantitative results of different methods.
Underline indicates the best performance among all methods.
Bold indicates the best performance among all non-fully supervised methods.
 `$E$(K)' denotes EVS-Net with `K' labeled patches. The `ALSSS/ $500$' and `USSS/ $500$' denote methods with $10$ or $500$ labeled patches (All scores are the average of three runs). }
\label{comparing_SOTA}
\end{table*}

\begin{figure*}[!t]
\centering
\vspace{-0.2em}
\includegraphics[scale =0.52]{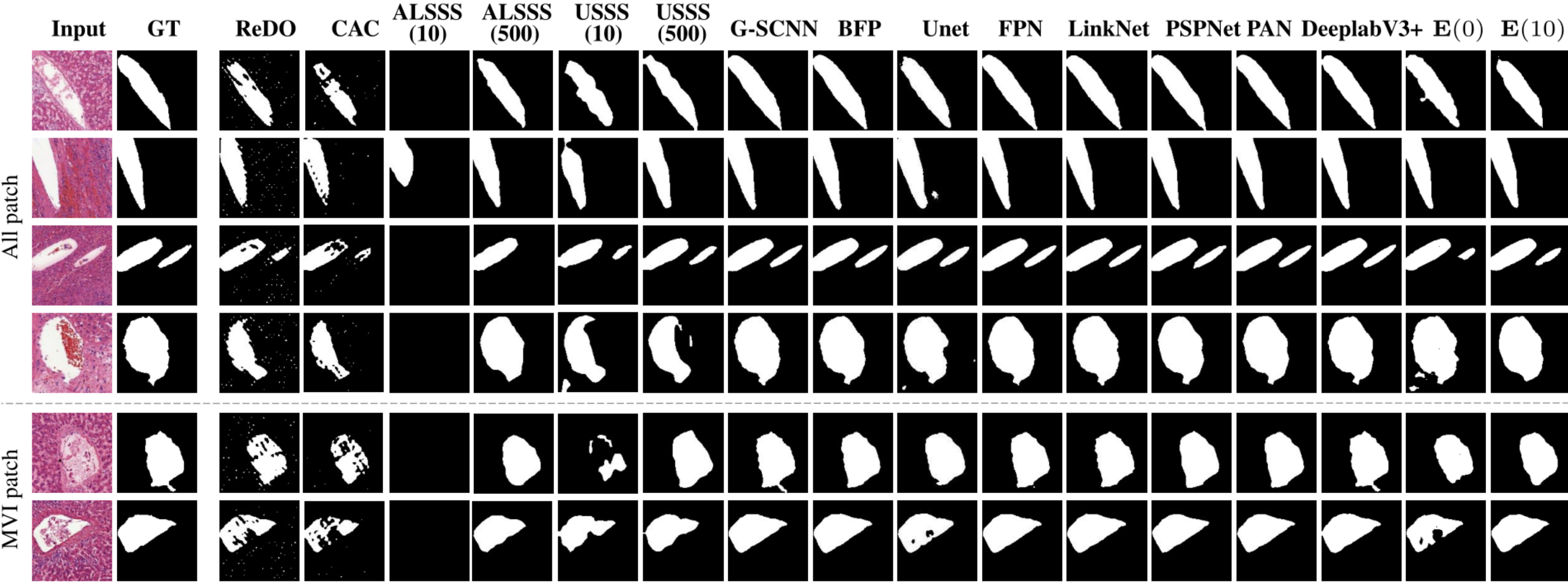}
\vspace{-0.4em}
\caption{The visual result of different methods. The `$E(K)$' denotes the method trained with $K$ labeled patches.
}
\vspace{-1.4em}
\label{fig:visualresults}
\end{figure*}

\subsection{Quantitative Evaluation}

\begin{figure*}[!t]
\centering
\vspace{-1.0em}
\includegraphics[scale =0.525]{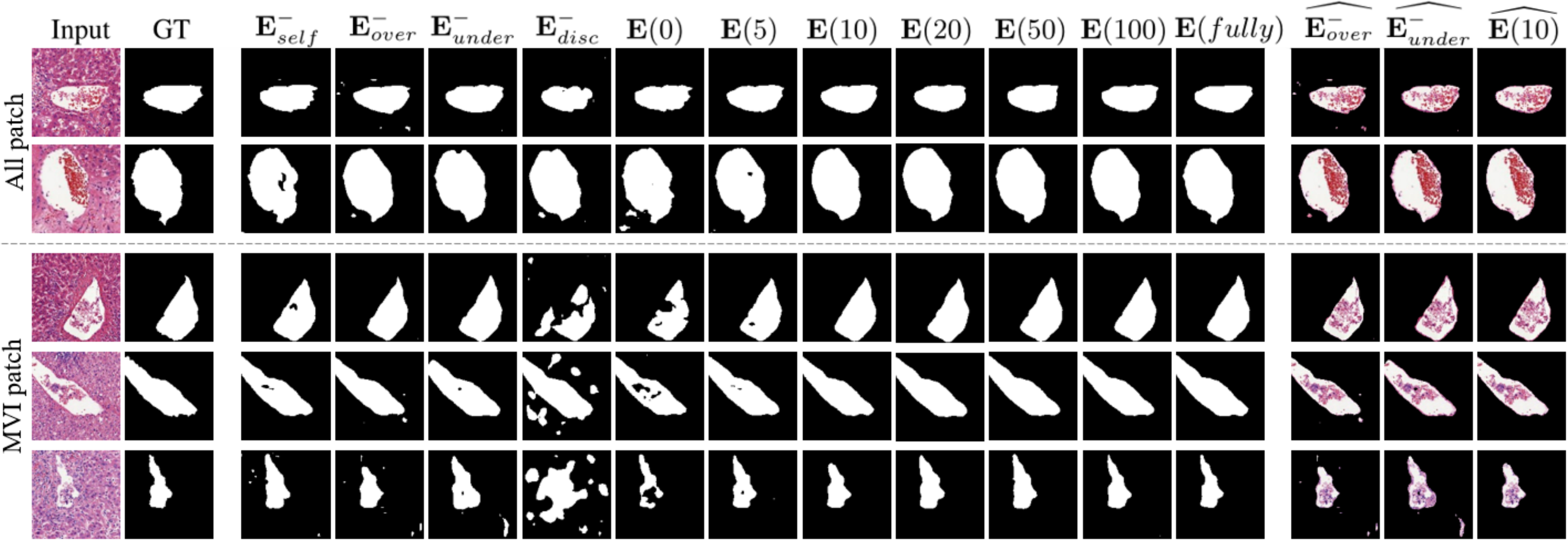}
\vspace{-1.6em}
\caption{The visual result of ablation study.
$\widehat{E}$ denotes segmented vessels of method $E$ (Zooming in to compare edges).}
\vspace{-1.5em}
\label{fig:ablationStudy}
\end{figure*}

To verify the effectiveness of EVS-Net, we compare the proposed method with the SOTA methods, including \emph{unsupervised methods} (ReDO~\cite{chen2019unsupervised}, CAC~\cite{hsu2018co-attention}),
\emph{weakly-/semi-supervised methods} (ALSSS~\cite{hung2018adversarial}, USSS~\cite{kalluri2019universal}) and
\emph{fully supervised methods } (edge-aware methods:\{ G-SCNN~\cite{takikawa2019gated-scnn:}, BFP~\cite{ding2019boundary-aware}\},Unet~\cite{ronneberger2015u-net:}, FPN~\cite{lin2017feature}, PSPNet~\cite{zhao2017pyramid}, PAN~\cite{li2018pyramid} and DeeplabV3+~\cite{chen2017rethinking}).
For the weakly-/semi-supervised methods (USSS and ALSSS), we provide two groups of labeled patches ($10$ patches and $500$ patches).
The same data augmentation strategy is applied on the above labeled patches.
Table~\ref{comparing_SOTA} shows the quantitative results, where all the scores are average of three runs.
For each trained model, we test it on all patches (`All patch') and patches ('MVI patch') that contain hepatocellular carcinoma cells.
It is obvious that DeeplabV3+ achieves the best performance among all the methods.
The $\mathbf{E}(100)$ achieves SOTA results on par with existing non-fully supervised methods.
Even with $10$ labeled samples,  \emph{EVS-Net} still achieves higher scores than non-fully supervised methods trained with hundreds of labeled patches on the `All patch'.
In addition, even without labeled samples,  $\mathbf{E}(0)$ (given in Table \ref{ablation_table}) still achieves better results than weakly/semi-supervised methods trained with $10$ labeled patches, which validates the effectiveness of the edge-competing mechanism.
 For `All patch', $\mathbf{E}(10)$ has about $2\%$ increase on the scores of $\mathbf{E}(0)$.
Meanwhile, for `MVI patch', $\mathbf{E}(10)$ has about $4\%$ performance increase compared with $\mathbf{E}(0)$,
which verifies the necessity of the limited labeled samples. More results of EVS-Net with different labeled patches are given in Table~\ref{ablation_table} and Fig.~\ref{fig:ablationStudy}.

\begin{table*}[!t]
\resizebox{\textwidth}{!}{
\centering
\begin{tabular}{c|cccccccccccc}
\toprule
&\multicolumn{1}{c}{\textbf{Index}$\backslash$\textbf{Ablation}}
& $\mathbf{E}_{self}^{-}$
& $\mathbf{E}_{over}^{-}$
& $\mathbf{E}_{under}^{-}$
& $\mathbf{E}_{disc}^{-}$
& $\mathbf{E}(0)$
& $\mathbf{E}(5)$
& $\mathbf{E}(10)$
& $\mathbf{E}(20)$
& $\mathbf{E}(50)$
& $\mathbf{E}(100)$
& $\mathbf{E}(fully)$ \\
 \cmidrule(r){1-2} \cmidrule(r){3-6} \cmidrule(r){7-13}
\multirow{4}{*}{\rotatebox{90}{All patch}}
&MPA    &95.73  &94.28  &93.59  &90.90  &94.77  &95.47  &\emph{96.24}  &96.38  &\underline{96.72}  &96.59  &\textbf{98.48}   \\
&MIoU   &92.38  &90.69  &90.02  &78.83  &91.93  &92.40  &\emph{93.34}  &93.55  &93.68  &\underline{94.38}  &\textbf{97.25}   \\
&FWIoU  &95.44  &94.02  &93.63  &86.38  &95.20  &95.46  &\emph{96.02}  &96.21  &96.22  &\underline{96.67}  &\textbf{98.37}   \\
&DICE   &98.56  &97.42  &97.30  &95.08  &98.50  &98.58  &\emph{98.75}  &98.76  &98.81  &\underline{98.96}  &\textbf{99.50}   \\
\cdashline{1-2}[0.8pt/2pt] \cdashline{3-6}[0.8pt/2pt] \cdashline{7-13}[0.8pt/2pt]
\multirow{4}{*}{\rotatebox{90}{MVI patch}}
&MPA    &87.80  &88.42  &87.26  &80.47  &85.12  &87.89  &\emph{89.81}  & 89.63 &89.25  &\underline{90.85}  &\textbf{97.16}   \\
&MIoU   &78.97  &78.73  &78.05  &62.22  &77.57  &81.32  &\emph{82.03}  & 82.91 &83.14  &\underline{86.19}  &\textbf{95.05}   \\
&FWIoU  &87.33  &86.64  &86.32  &74.16  &86.74  &89.01  &\emph{89.25}  & 89.83  &90.10  &\underline{91.98}  &\textbf{97.14}   \\
&DICE   &95.71  &94.75  &94.67  &89.19  &94.66  &96.43  &\emph{96.42}  & 96.35 &96.80  &\underline{97.46}  &\textbf{99.12}   \\
\bottomrule
\end{tabular}
}
\vspace{-0.5em}
\caption{The ablation study result on EVS-Net.
$E_{self}^{-}$, $E_{cond}^{-}$, $E_{over}^{-}$, $E_{under}^{-}$ and  $E_{disc}^{-}$ denote the EVS-Net without edge-aware self-supervision, over-segmentation discriminator, under-segmentation discriminator, and two discriminators.
$E(K)$ denotes the EVS-Net with $K$ labeled patches.
Italic indicates the best performance among methods trained with $10$ labeled patches.
Bold and underline indicate the best and second-best performance among all methods (All scores in $\%$).}
\label{ablation_table}
\end{table*}

\subsection{Qualitative Evaluation}

The visual results of different methods are shown in Fig.~\ref{fig:visualresults},
where we can see that most of the fully supervised methods achieve similar results.
Deep learning-based methods indeed have the potential to be applied to the pathological liver vessel segmentation.
For the unsupervised methods (ReDO and CAC), the segmented background and vessels contain many small holes, which indicates the instability of the unsupervised methods.
From the column $5$, we can see that ALSSS with $10$ labeled patches fails to segment most cases, demonstrating that ten labeled patches cannot support enough supervision information for the semi-supervised method. Even with $500$ labeled patches, USSS still fails to segment the vessel with a large hepatocellular carcinoma area (row 4, column 8).
For the blurry edge vessels (row 1 and row 5), EVS-Net with larger than $10$ labeled patches can achieve more accurate edge segmentation than non-fully supervised methods.
For MVI patches, $\mathbf{E}(0)$ achieves inaccurate segmentation on those hepatocellular carcinoma areas,
which indicates that limited labeled patches are necessity for guiding the segmentation of hepatocellular carcinoma areas.
In summary, with limited labeled patches, EVS-Net can achieve close results on par with fully supervised methods.

\subsection{Ablation Study}
To verify the effectiveness of each component,
 we conduct an ablation study on the edge-aware self-supervision, over-segmentation discriminator, under-segmentation discriminator, two discriminators, and the different numbers of labeled patches.
 The $\mathbf{E}_{self}^{-}$, $\mathbf{E}_{over}^{-}$, $\mathbf{E}_{under}^{-}$ and  $\mathbf{E}_{disc}^{-}$ denote the EVS-Net without edge-aware self-supervision, over-segmentation discriminator, under-segmentation discriminator, and two discriminators,
 which are trained in the same setting of $10$ labeled patches.
 Table~\ref{ablation_table} shows the quantitative results of the above methods.
  The $\mathbf{E}(K)$ denotes the EVS-Net with $K$ labeled patches.
 We can see that $\mathbf{E}(10)$ achieves the best performance among $\mathbf{E}_{self}^{-}$, $\mathbf{E}_{over}^{-}$, $\mathbf{E}_{under}^{-}$ and  $\mathbf{E}_{disc}^{-}$, which demonstrates that all components of EVS-Net are helpful for the segmentation performance.
 The performance of $\mathbf{E}_{disc}^{-}$ that doesn't contain two discriminators degrades about $10\%$ compared with the $\mathbf{E}(10)$.
 Methods with only one discriminator ($\mathbf{E}_{over}^{-}$ and $\mathbf{E}_{under}^{-}$) reduce about $2\%$ compared with the $\mathbf{E}(10)$. From Fig.~\ref{fig:ablationStudy}, we can see that predicted masks of  $\mathbf{E}(10)$ have accurate edges than $\mathbf{E}_{over}^{-}$ and $\mathbf{E}_{under}^{-}$, which verifies the effectiveness and necessity of edge-competing.
 In addition, Fig.~\ref{fig:ablationStudy} shows that the result of $\mathbf{E}_{self}^{-}$ has some holes.
 By contrast, the result of $\mathbf{E}(10)$ is accurate, which indicates that the edge-aware self-supervision is also beneficial for eliminating the incorrect holes.
For the different numbers of labeled patches, we find that the critical cut-off point is $10$-labeled-samples, which can supply relatively sufficient guidance. For the patches that have MVI, vessels' edges are usually blurry, and vessels' inner usually contain the large area of the hepatocellular carcinoma cells.
Without any guidance of labeled samples, the pathological vessel segmentation network will fail to obtain the vessels' desired edges.
Therefore, the $\mathbf{E}(0)$ achieves the worst performance among all $\mathbf{E}(K)$ methods, which also verifies the necessity of the limited labeled patches.
In addition, the comparisons between the segmented vessels of $\mathbf{E}_{over}^{-}$, $\mathbf{E}_{under}^{-}$, $\mathbf{E}(10)$ are shown in
 the last three columns on the right of Fig.~\ref{fig:ablationStudy},
 where we can see that segmented vessels'  contours of $\mathbf{E}_{over}^{-}$ and $\mathbf{E}_{under}^{-}$ contract inward and expand outward, respectively. With the edge-competing between OS-D and US-D, $\mathbf{E}(10)$ have accurate edges' contours, which effectively clarifies the effectiveness of the edge-competing mechanism.

\section{Conclusion}
In this paper, we propose the first computer-aided diagnosis method for the pathological liver vessel segmentation,
which is an essential and critical step for the diagnosis of MVI.
For massive unlabeled pathological patches,
the doctor only needs to label limited patches.
Next, binarization processing combining with erosion and dilation is adopted to generate some under-segmented and under-segmented masks.
Then, the EVS-Net is proposed for segmenting the pathological liver vessel with limited labels.
The EVS-Net contains a pathological vessel segmentation network and two discriminators.
With the limited labeled patches, the segmentation network is initialized with the supervision training.
Meanwhile, an edge-aware self-supervision module is proposed for enhancing the edge consistency of massive unlabeled patches.
With the over-segmented and under-segmented patches,
two discriminators are devised for distinguishing
whether the segmented vessels and background contain residual features.
The segmentation network and two discriminators are trained in an adversarial manner.
The edge-competing mechanism is verified to be very effective for facilitating the segmentation network segmenting more accurately.
Exhaustive experiments demonstrate that the proposed EVS-Net achieves a close performance of fully supervised methods with limited labeled patches.
It provides a convenient tool for  the pathological liver vessel segmentation.
In future work, we will focus on detecting and counting the hepatocellular carcinoma cells in the segmented vessels,
which will provide a complete assisting diagnosis method for the diagnosis of MVI.

\section{ Acknowledgments}
This work was supported by National Natural Science Foundation of China (No.62002318), Zhejiang Provincial Natural Science Foundation of China (LQ21F020003), Zhejiang Provincial Science and Technology Project for Public Welfare (LGF21F020020), Programs Supported by Ningbo Natural Science Foundation (202003N4318), and Alibaba-Zhejiang University Joint Research Institute of Frontier Technologies.
\bibliography{MYRE}

\end{document}